\documentclass[runningheads]{llncs}
\usepackage{graphicx}
\usepackage{booktabs}
\usepackage{amsmath}
\usepackage{subcaption}
\begin{document}
\title{G-SciEdBERT: A Contextualized LLM for Science Education in German}

\titlerunning{G-SciEdBERT: A Contextualized Large Language Model in German}
%
\author{Ehsan Latif\inst{1} \and
Gyeong-Geon Lee\inst{1} \and 
Knut Neumann\inst{2} \and
Tamara Kastorff\inst{3} \and
Xiaoming Zhai\inst{1}\thanks{Corresponding author: xiaoming.zhai@uga.edu}
}
\authorrunning{Latif et al.}
%
\institute{AI4STEM Education Center \& Department of Mathematics, Science, and Social Studies Education, University of Georgia, Athens, GA, USA
\email{ehsan.latif,ggleeinga,xiaoming.zhai}@uga.edu\and
IPN–Leibniz-Institute for Science and Mathematics Education, Kiel, Germany\\
\email{neumann@leibniz-ipn.de}\and
School of Social Sciences and Technology, Centre for International Student Assessment (ZIB), Technical University of Munich, Germany\\
\email{tamara.kastorff@tum.de}}

\maketitle              
\begin{abstract}
The advancement of natural language processing has paved the way for automated scoring systems in various languages, such as German (e.g., German BERT [G-BERT]). Automatically scoring written responses to science questions in German is a complex task and challenging for standard G-BERT as they lack contextual knowledge in the science domain and may be unaligned with student writing styles. This paper presents a contextualized German Science Education BERT (G-SciEdBERT), an innovative large language model tailored for scoring German-written responses to science tasks and beyond. Using G-BERT, we pre-trained G-SciEdBERT on a corpus of 30K German written science responses with 3M tokens on the Programme for International Student Assessment (PISA) 2018. We fine-tuned G-SciEdBERT on an additional 20K student-written responses with 2M tokens and examined the scoring accuracy. We then compared its scoring performance with G-BERT. Our findings revealed a substantial improvement in scoring accuracy with G-SciEdBERT, demonstrating a 10.2\% increase of quadratic weighted Kappa compared to G-BERT (mean difference = 0.1026, SD = 0.069). These insights underline the significance of specialized language models like G-SciEdBERT, which is trained to enhance the accuracy of contextualized automated scoring, offering a substantial\textit{} contribution to the field of AI in education.

\end{abstract}
\section{Introduction}
Automatic scoring systems have significantly evolved, leveraging advanced machine-learning techniques to offer accurate and efficient scoring for student responses. Studies have systematically reviewed the application of machine learning in science assessment \cite{zhai2020applying}, highlighting the growing promise of Artificial Intelligence (AI) to support assessment practices \cite{zhai2022applying}.


The characteristics of language and subject matter play crucial roles in automatic scoring systems using language models \cite{alsanie2022automatic}
To address this issue, global scholars have strived to develop language- and/or domain-specific language models. For example, to adapt to the needs of Germany, researchers developed German BERT (G-BERT), which has been applied in various tasks  \cite{fernandez2022automated}
including automatic scoring of reading comprehension, spontaneous speech, 
written essays \cite{sawatzki2021deep,ludwig2021automated}, comprehension for medical domain \cite{bressem2024medbert}, and content summarization \cite{iskender2020best}. A recent study \cite{gururangan2020don} also highlights the significance of pre-training to achieve high gains for domain-specific tasks. Despite this progress, little research has been conducted in specific fields like German science education, which is grounded in a unique educational context. Whereas there are prominent specific large language models (LLMs) (i.e., G-BERT), they have not been practically served for science education and science assessment specifically—an endeavor that poses special difficulties because of its intricate and sophisticated concepts, terminologies, syntax, and semantics.

To fill these gaps, this study aims to create a contextually learned model, especially one that is pre-trained for German science education, called German Science Education BERT (G-SciEdBERT). We pre-trained G-SciEdBERT based on the existing G-BERT using more than 30k student written responses from the Programme for International Student Assessment (PISA) dataset and an additional 20K responses used for fine-tuning.
Different from its predecessors, G-SciEdBERT is specially developed for students who are navigating the complexities of scientific language for written science tasks. Rather, G-SciEdBERT tackles the shortcomings of domain-general LLMs by including domain-specific information in its training process, offering a more accurate and nuanced natural language processing of student writing in science education.

The impact of G-SciEdBERT on the complex requirements of science assessment is just as important as its specialized pre-training. This model exemplifies how subject-specific AI capabilities advance autonomous scoring systems applicable to global educational situations. The nexus between AI and science education will be a rich source of innovation as we learn more about the features and applications of G-SciEdBERT, a language- and domain-specific LLM.

In the following, we highlight the major contributions of the paper:
\begin{itemize}
    \item We built a contextualized Large Language Model termed G-SciEdBERT by following the pre-training BERT pipeline. G-SciEdBERT was developed based on contextual learning ideas with a carefully designed input format that captures the context of the subject domains and the written features of secondary students.
    \item We evaluated the performance of our G-SciEdBERT through experimental validation with student responses to German science assessment tasks. Our approach substantially outperforms state-of-the-art G-BERT \cite{chan2020german}.
    \item We have open-sourced the code at GitHub\footnote{\url{https://github.com/ehsanlatif/G-SciEdBERT.git}} for reproducibility and released the final version of G-SciEdBERT to Hugging face\footnote{\url{https://huggingface.co/ai4stem-uga/G-SciEdBERT}} for community to use.
\end{itemize}

\section{Background}
\subsection{Contextualized Large Language Models}

The recent advent of LLMs like BERT \cite{devlin2018bert} and GPT \cite{achiam2023gpt} have ushered in a new era of automatic scoring capabilities \cite{kasneci2023chatgpt,2023Lee}, offering nuanced understanding and evaluation of text response \cite{grassini2023shaping}. Most LLMs are trained on publically available academic and general writings, which has shown fewer advantages for automatic scoring of student-written responses in a specific subject domain \cite{2023Wu}. To improve the scoring performance, researchers have to fine-tune LLMs. For instance, Latif and Zhai \cite{latif2023fine} fine-tuned GPT-3.5 Turbo and reported high automatic scoring accuracies for complex written responses to science tasks. Parallel studies also reported how fine-tuned LLMs have significantly improved scoring accuracy for science \cite{condor2021automatic,lee2023gemini}, mathematics \cite{erickson2020automated,shen2021mathbert}, and essay writing \cite{mayfield2020should}. However, these fine-tuned LLMs are not without their limitations, particularly in handling diverse languages with robust performance across datasets. One solution is to further pre-train LLMs in context (e.g., SciEdBERT)\cite{liu2023context}, which has shown significant improvement compared to the general LLMs. To address the needs of other languages, researchers have shown improved multilingual automatic scoring capabilities of BERT \cite{horbach2023crosslingual,nozza2020mask,2023Gupta} using optimized pre-training \cite{liu2019roberta}.

\subsection{Cross-Prompt Automatic Scoring}
The task of building contextualized LLMs is usually classified as cross-prompt automatic scoring, which is a major research domain, especially in the field of automated essay scoring (AES). The cross-prompt technique is especially pertinent for scalable educational applications, as it allows for a single model to accurately score essays or written responses across different topics or domains without retraining for each specific prompt. Our work on G-SciEdBERT operates within this emerging research landscape, aligning with and extending the current methodologies in several key areas.

Cross-prompt scoring methodologies have been developed to address the challenges posed by varying essay prompts. A study introduced a domain generalization approach for cross-prompt AES, emphasizing the need for models capable of adapting to unseen prompts \cite{ridley2020prompt}. Expanding on this, authors further explored automated cross-prompt scoring of essay traits, presenting techniques that mitigate the variability in essay content and structure across different prompts \cite{ridley2021automated}. These foundational studies underscore the significance of prompt adaptability in AES systems.

The prompt-map AES which leverages prompt-mapping contrastive learning to enhance cross-prompt scoring accuracy \cite{chen2023pmaes}. This method aims to bridge the gap between prompts by identifying and aligning latent semantic structures shared across essays responding to different questions. Similarly, researchers \cite{funayama2023can,funayama2023reducing} also explored the benefits of cross-prompt training data for short answer scoring models, suggesting that leveraging diverse training data can improve model robustness and performance across various prompts.

Studies also applied Sentence-BERT \cite{sasaki2022sentence} to distinguish between high and low-quality essays in a cross-prompt setting, highlighting the utility of sentence embeddings in capturing nuanced differences in writing quality irrespective of the prompt. This approach aligns with our use of BERT-based models, though our work specifically tailors the underlying architecture and training data to the context of German science education.

Another study \cite{song2020multi} discussed multi-stage pre-training for automated Chinese essay scoring, a technique that could parallel our method of pre-training G-SciEdBERT on a specialized corpus before fine-tuning for assessment tasks. Their work emphasizes the importance of domain-specific pre-training, a principle that underpins our approach to developing a contextually aware model for the German science education sector.

Building upon these foundations, our G-SciEdBERT model introduces a novel application of contextualized large language models for cross-prompt scoring in the specific domain of science education. By pre-training on a domain-specific corpus of German science education texts, G-SciEdBERT is designed to understand and evaluate the complex, terminology-rich responses characteristic of science assessments. This focus on domain specificity and contextual learning sets our work apart from existing general-purpose and cross-prompt AES systems, providing a tailored solution for the nuanced challenges of scoring science education responses.

While the field of cross-prompt AES has progressed through various innovative approaches, G-SciEdBERT represents a pioneering step toward domain-specific, contextually aware scoring models. Our model not only contributes to the body of research on cross-prompt scoring but also opens new avenues for the application of LLMs in educational assessment, particularly within the German language context. In this study, we aim to analyze the effect of contextualization on LLM for automatic scoring through per-training.


\section{Method}
In this study, we aim to pre-train and fine-tune the BERT model for the given complex German written scientific response. Here, we first define the problem statement, discuss the dataset used, and explain the approach to build G-SciEdBERT, including pre-training and fine-tuning. 

\subsection{Problem Formulation}

In the context of educational assessments, we have a set of scientific tasks $T_i \in \mathcal{T}_{1,2,...,n}$. The assessment process involves evaluating student responses against a scoring rubric that provides detailed criteria for assigning scores based on various levels of understanding and completeness.

The primary components associated with each assessment item are:
\begin{enumerate}
    \item \textbf{Scientific Task $ T_i $}: This is the central challenge or question posed to the students, designed to test their understanding and application of scientific concepts.
    \item \textbf{Scoring Rubric}: A detailed framework that defines the criteria for evaluating responses. The rubric categorizes responses into different levels, such as correct, partially correct, and incorrect, based on the presence of key concepts, reasoning, and completeness.
    \item \textbf{Training Dataset $ D^{\text{train}}_i $}: This dataset contains a large number of human-scored student responses written in German. Each response $ x_i $ in the dataset is paired with a score $ y_i $, where $ y_i $ reflects the assessment of the response according to the scoring rubric.
\end{enumerate}

The training dataset can be formally represented as $ D^{\text{train}}_i = \{(x_i, y_i)\} $, where:
\begin{itemize}
    \item $ x_i $ is a vector representing the textual response of a student to the task $ T_i $.
    \item $ y_i $ is the corresponding score or label assigned to the response $ x_i $, which may be a categorical value (e.g., excellent, good, fair, poor) or a numerical score (e.g., 1 to 5).
\end{itemize}

The goal is to develop models capable of automatically predicting the scores for new, unseen student responses in the test dataset $ D^{\text{test}}_i = \{(x^{\text{target}}_i)\} $. The test dataset consists of responses $ x^{\text{target}}_i $ without associated scores, representing the real-world application where the model's predictions are used to evaluate student performance.


Our approach addresses this by leveraging the shared semantics across items. Instead of training individual models for each task, we pre-train a single shared model, named G-SciEdBERT, using the G-BERT architecture. This model is designed to capture and utilize the contextual information from multiple tasks, enabling it to generalize better across different items and improve scoring accuracy. By integrating shared scientific content knowledge, G-SciEdBERT aims to enhance the robustness and fairness of automated scoring systems in educational settings.

\subsection{Dataset Details}
PISA is an international large-scale assessment led by the Organisation for Economic Co-operation and Development (OECD) that has been carried out every three years since 2001. Historically, PISA assessed the mathematical, scientific, and reading literacy of 15-year-old students in participating countries. This study used data from the German PISA 2015 and 2018 samples. In 2015, the sample included a total of 6,522 students from 256 schools \cite{pisa2015results}; in 2018, the sample included a total of 5,451 students from 223 schools \cite{pisa2018results}. Due to the design of the PISA study, the samples represent the general German student population.

Our analysis focused on 59 constructed response items from PISA 2015, 32 of which were also run in PISA 2018. The items were either short responses (around one sentence) or extended response items (up to five sentences). Student responses had 20 words on average. The possible scores ranged from 0 to 5, with the maximum being 3 or 4 for short responses and 4 or 5 for extended responses. Student responses had been scored as part of the respective iteration of PISA. In PISA 2015, student responses were scored by eight coders. For this purpose, items were grouped into four sets. Each coder was assigned to two item sets, such that coding of each item set was performed by four coders. For coding, student responses were split evenly across coders; that is, each coder coded one-fourth of the responses in a given item set.

Every coder coded a total of 100 randomly selected responses to each item. Coder agreement was 93.4 percent for trend items (i.e., items that had been part of a previous iteration) and 92.3 percent for new items (i.e., items administered the first time in PISA 2015). Not a single item had an agreement below 85 percent \cite{pisa2015techbook}. In PISA 2018, student responses by four to five coders were due to the lower number of items (in PISA 2015, scientific literacy was the focus and hence assessed using more items). Again, a total of 100 randomly selected responses were coded by all coders. The agreement was 96.4 percent (on the 32 trend items already used in PISA 2015) \cite{pisa2018techbook}. Responses were coded per item in both iterations, irrespective of the student's ethnicity, race, or gender, to ensure fairness.

We utilized student responses to the 32 items from both iterations of PISA to pre-train the G-SciEdBERT. Student responses to the remaining 27 items from PISA 2015 were utilized for fine-tuning and automatic scoring. Our pre-training dataset contains more than 30,000 student-written German responses, which means approximately 1,000 human-scored student responses per item for contextual learning through fine-tuning.

\subsection{Our Approach: G-SciEdBERT}

\textbf{G-SciEdBERT Pre-training:} 

Pre-training is a crucial step in developing G-SciEdBERT. It enables the model to learn a robust representation of the German language and scientific content, which is essential for accurately scoring student responses. The pre-training phase leverages a large corpus of text to train a general language model, which can then be fine-tuned on a specific downstream task—in this case, the automatic scoring of German-written scientific responses.

G-BERT, the foundational model used in this process, is a BERT-based model pre-trained on extensive German language corpora\footnote{G-BERT \cite{chan2020german} is a specialized BERT model trained on 12GB German textual content (details can be found at \url{https://www.deepset.ai/german-bert}) is the most suitable candidate for our training purpose because of it high masking accuracy (0.905 for 10kGNAD dataset \cite{scheible2020gottbert})}. It consists of 110 million parameters and employs a masked language model (MLM) objective. This objective involves predicting masked words in a sentence, helping the model learn the context and semantics of the language. While G-BERT is proficient in general German language understanding, it lacks the specialized training required for scoring scientific responses. It involves understanding not just language but also scientific assessments' specific content and context.

Although effective for general language tasks, G-BERT does not possess the specialized knowledge required to accurately score student responses in scientific assessments. This limitation arises because G-BERT is not pre-trained on domain-specific data relevant to educational assessments or scientific content. Consequently, while G-BERT can understand general German text, it may not accurately interpret the nuances and specific knowledge required to assess scientific responses, such as understanding scientific concepts, terminologies, and the logical structure of answers.

To bridge this gap, G-SciEdBERT is pre-trained using a domain-specific corpus of Student-Written scientific responses. This corpus includes 30,000 responses, comprising approximately 3 million tokens, which provide rich contextual information about how students articulate scientific content. The pre-training process involves learning the general German language and the specific vocabulary and common answer structures found in scientific assessments.

During pre-training, G-SciEdBERT is exposed to input sequences that combine the scientific task description $ T_i $ and the student response $ x^{\text{target}}_i $. These segments are separated by the [SEP] token, which helps the model differentiate between the task context and the response content. This structured input allows the model to learn to associate certain types of responses with particular tasks and to understand how different answers should be scored based on the context provided by the task description.

A special handling mechanism is employed because scientific responses can often exceed the maximum input length of 512 tokens set by G-BERT. The text is first encoded using a fixed, non-trainable G-BERT model, and the resulting [CLS] embedding vector, which represents a summary of the entire input sequence, is then passed to G-SciEdBERT. This vector is crucial because it captures the essence of the response and the associated task context, enabling the model to make more informed scoring decisions.

Finally, the [CLS] embedding vector is fed into a linear classification layer, which outputs the score prediction. A softmax function facilitates the transformation from embeddings to score predictions \cite{goodfellow2016deep}:

\begin{equation}
    P(y|x) = \text{softmax}(W \cdot \text{[CLS]} + b)
\end{equation}

where $ W $ is the weight matrix and $ b $ is the bias term. This final layer maps the contextualized embeddings to the scoring labels, which can range from numerical scores to categorical labels such as "excellent," "good," "fair," and "poor."

The pre-training process significantly enhances the model's ability to understand and assess student responses in a scientifically accurate manner, making G-SciEdBERT a specialized tool for automatic scoring in educational assessments. This adaptation from general language understanding to domain-specific expertise is what differentiates G-SciEdBERT from its predecessor, G-BERT.

\begin{figure}
    \centering
    \includegraphics[width=1\linewidth]{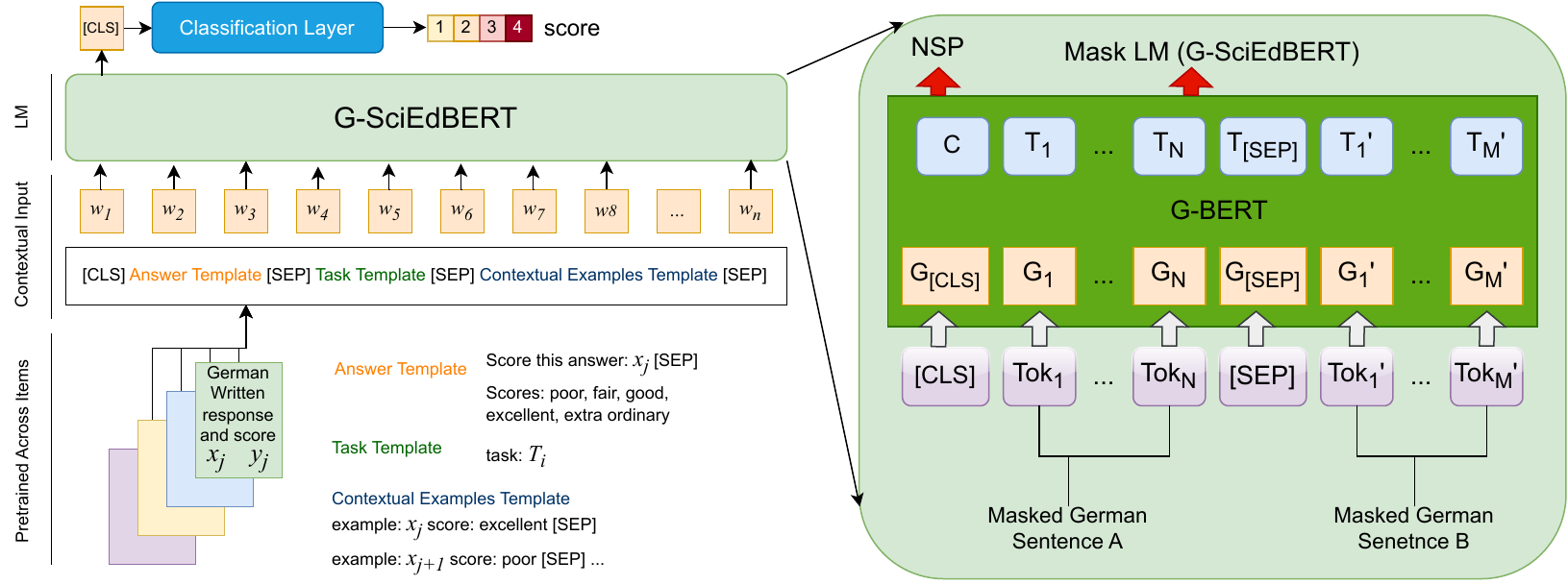}
    \caption{System Architecture: G-SciEdBERT pretraining and fine-tuning to score German written responses automatically.}
    \label{fig:architecture}
\end{figure}

\textbf{G-SciEdBERT Fine-Tuning:}

Fine-tuning is another essential process in the development of G-SciEdBERT, transforming the pre-trained model into a specialized tool for automatically scoring scientific responses. While pre-training equips the model with a broad understanding of the language and general concepts, fine-tuning tailors this knowledge to the specific task of scoring student responses based on detailed rubrics and criteria.

During fine-tuning, G-SciEdBERT is exposed to a domain-specific dataset, such as the PISA 2015 dataset, including students' scored responses. This phase is crucial because it adjusts the model's pre-learned parameters to align more closely with the specifics of the scoring task, such as recognizing the importance of certain scientific concepts, the clarity of explanation, and the logical flow in responses.

The fine-tuning process significantly enhances the model's ability to make accurate predictions by minimizing the cross-entropy loss between the predicted and true scores provided by human graders. This loss function measures the difference between the model's output and the actual labels, allowing the model to learn and correct its predictions. The reduced loss translates to improved accuracy in automated scoring, ensuring that the model's decisions align closely with human judgment.

The fine-tuning of G-SciEdBERT involves feeding the model with a structured input format that combines three essential components:

\begin{enumerate}
    \item  This template prompts the model to focus on the specific student response $ x_j $ that needs to be scored. The template often includes directives like "Score this answer: $ x_j $" followed by possible score categories, which may range from "poor" to "extraordinary." This setup helps the model frame the response in the scoring context, directing attention to key features relevant to the evaluation.
    \item By including the specific task $ T_i $, the model receives contextual information that grounds the scoring process in the details of the question being answered. This template helps ensure that the scoring considers the specific requirements and expectations associated with the scientific task, which might include understanding certain principles or applying specific methodologies.
    \item This component provides the model with examples of other responses and their corresponding scores. For instance, "example: $ x_j $, score: excellent" and "example: $ x_{j+1} $, score: poor" serve as reference points. These examples help the model understand the scoring rubric better by illustrating what constitutes a high-quality response versus a lower-quality one. This comparative context enhances the model's ability to discriminate between different levels of response quality.
\end{enumerate}

The fine-tuning process allows G-SciEdBERT to specialize in the task of scoring scientific responses, which involves not only linguistic understanding but also domain-specific knowledge and the application of complex scoring criteria. This process refines the model's internal representations. It improves its ability to generalize from training data to unseen test data, thereby increasing the reliability and fairness of the automated scoring system.

By fine-tuning a dataset similar to the test data but separate from the data used during pre-training, the model becomes adept at handling variations in student responses and can better cope with the diversity of real-world data. This separation ensures that the model's performance is robust and not overly reliant on any specific dataset's peculiarities, enhancing its generalization capabilities.

The result is a powerful scoring model that combines the broad linguistic capabilities learned during pre-training with the detailed, task-specific insights gained during fine-tuning. This dual-phase training strategy is essential for developing a model that can accurately and consistently score student responses in an automated and scalable manner.

For reproducibility and further adaptation, such as developing models for other languages or domains, detailed implementation steps, including pre-training and fine-tuning procedures, are documented in the GitHub\footnote{\url{https://github.com/ehsanlatif/G-SciEdBERT.git}}. This repository provides researchers with the necessary resources and guidance to replicate and extend the G-SciEdBERT approach.

The overall system architecture for G-SciEdBERT pretraining and fine-tuning for automatic scoring of German written responses is shown in Fig.~\ref{fig:architecture}.

\section{Experimental Evaluation}

\subsection{Metrics and Baseline}

In our experimental evaluation, we drew on quadratic weighted Kappa (QWK) as the primary metric for assessing the accuracy of the automatic scoring system. QWK is widely recognized in the field of educational assessment for its ability to measure agreement between two sets of categorical ratings, adjusting for chance agreement. It is particularly useful for evaluating the consistency between human scores and automated scoring systems.

The QWK was chosen because of its sensitivity to differences in scoring, as it penalizes both the magnitude and the direction of disagreements between predicted and actual scores. This characteristic makes QWK a more informative and robust metric than simple accuracy, especially in ordinal data like educational scores, where the distance between categories (e.g., scores) carries meaningful information. QWK provides a nuanced evaluation by considering the extent to which the automated scores deviate from the human scores, offering a more comprehensive assessment of the model's performance.

We used a paired \textit{t}-test for statistical comparison of the item-wise accuracy between G-SciEdBERT and the baseline model G-BERT \cite{chan2020german}. This test is appropriate for comparing the performance of the two models on the same items, allowing us to determine if the observed differences in their performance are statistically significant.

\subsection{Implementation Details}

\textbf{Preprocessing:}  
We applied automatic spelling correction using a German lexicon library to align the student responses with the data used during G-BERT's pre-training. This step was crucial for two reasons:
\begin{enumerate}
    \item The scoring rubrics explicitly instruct scorers to overlook minor spelling errors, focusing instead on content and understanding. Therefore, correcting these errors ensured that the model did not penalize students unfairly for spelling mistakes.
    \item Normalizing the textual data made the student responses more consistent with the text used in the pre-training phase, which helped the model leverage its pre-trained knowledge more effectively.
\end{enumerate}

\textbf{Experimental Setup:}  
We employed a five-fold cross-validation strategy to rigorously assess our model's performance. In this setup, the dataset was divided into five parts: three for training, one for validation, and one for testing. This method ensures that the model's evaluation is based on unseen data, providing a reliable estimate of its generalization capabilities.

All pre-training and fine-tuning implementations were carried out using the HuggingFace transformers library \cite{wolf2020transformers}, a widely used tool for NLP tasks. We utilized the Adam optimizer, with a batch size of 32 and a learning rate of $3 \times 10^{-5}$. The categorical cross-entropy loss function was employed, which is standard for classification tasks, as it measures the model's performance in predicting class probabilities.

In accordance with G-BERT's architecture, the maximum input length was set to 512 tokens. Importantly, we did not conduct hyperparameter tuning; instead, we adhered to the default settings provided by the G-BERT model due to its proven optimality for the German Language to maintain consistency and comparability.

The G-SciEdBERT model was first pre-trained using the PISA 2018 dataset, which provided a foundational understanding of the response types and scoring criteria. This pre-trained model was then fine-tuned on 27 different G-SciEdBERT models using the PISA 2015 dataset, deliberately excluding data already seen during the pre-training phase to avoid data leakage. The fine-tuning process ran for four epochs, as the model typically reached optimal performance on the validation set by the third epoch. Each epoch of G-SciEdBERT training required approximately 6 hours on a single NVIDIA RTX 8000 GPU, reflecting the computational intensity of the training process.

This comprehensive setup, combined with robust evaluation metrics, ensures that the results are both accurate and reliable, offering valuable insights into the performance of G-SciEdBERT in automatic scoring tasks.

For evaluation purposes, we compare the performance of fine-tuned G-BERT and G-SciEdBERT for automatic scoring of student-written science responses in the German language using PISA 2018 data. We have not included any prompting technique for this study, as the fine-tuned models are already contextualized for specific questions.

\begin{table}[h]
\centering
\caption{Comparison between G-BERT and G-SciEdBERT on automatic scoring of randomly selected five items from 27 PISA 2015 science items excluding responses of items from PISA 2018.}
\label{tab:accuracyComparison}
\begin{tabular}{@{}lcccccc@{}}
\toprule
& \multicolumn{3}{c}{ Samples (\textit{n}) } & \multicolumn{2}{c}{Accuracy as QWK} \\
Item      & Training Samples & Testing Samples & Labels & G-BERT & \textbf{G-SciEdBERT} \\ \midrule
S131Q02   & 487              & 122             & 5            & 0.761           & \textbf{0.852}                         \\
S131Q04   & 478              & 120             & 5            & 0.683           & \textbf{0.825}                       \\
S268Q02   & 446              & 112             & 2            & 0.757           & \textbf{0.893}                      \\
S269Q01   & 508              & 127             & 2            & 0.837           & \textbf{0.953}                    \\
S269Q03   & 500              & 126             & 4            & 0.702           & \textbf{0.802}                  \\
Average (27)  & 598             & 150             & 2-5 (min-max)            & 0.7758          & \textbf{0.8785}                       \\ \bottomrule
\end{tabular}
\end{table}

\begin{figure}
    \centering
    \begin{subfigure}[b]{\linewidth}
        \centering
        \includegraphics[width=\linewidth]{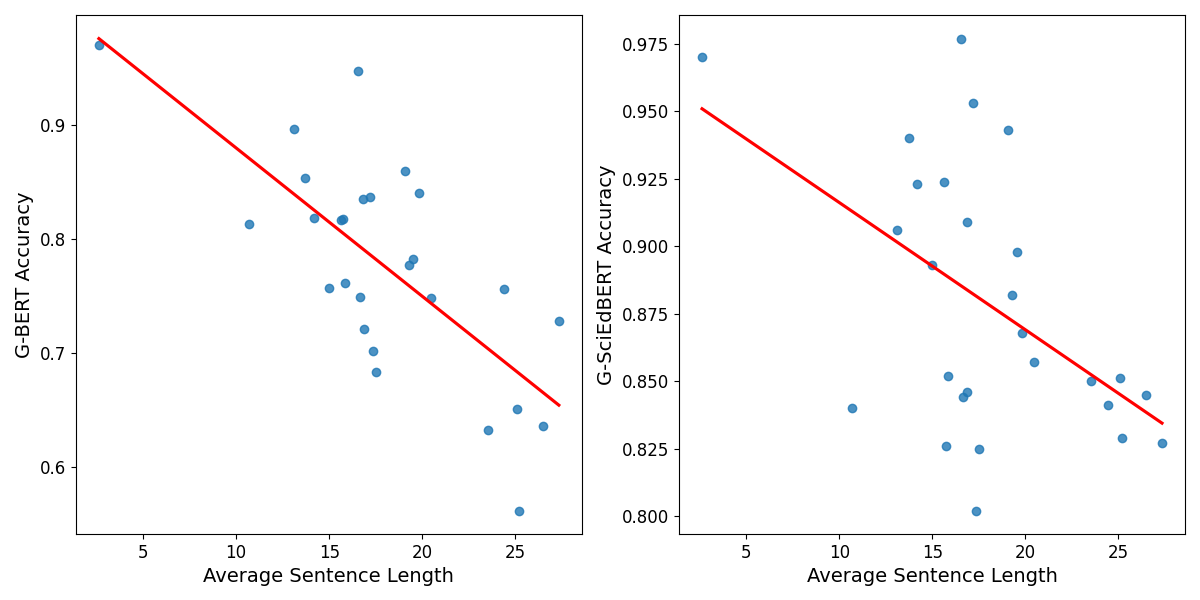}
        \caption{Effect of item-wise response length}
        \label{fig:length_vs_accuracy}
    \end{subfigure}
    \begin{subfigure}[b]{\linewidth}
        \centering
        \includegraphics[width=\linewidth]{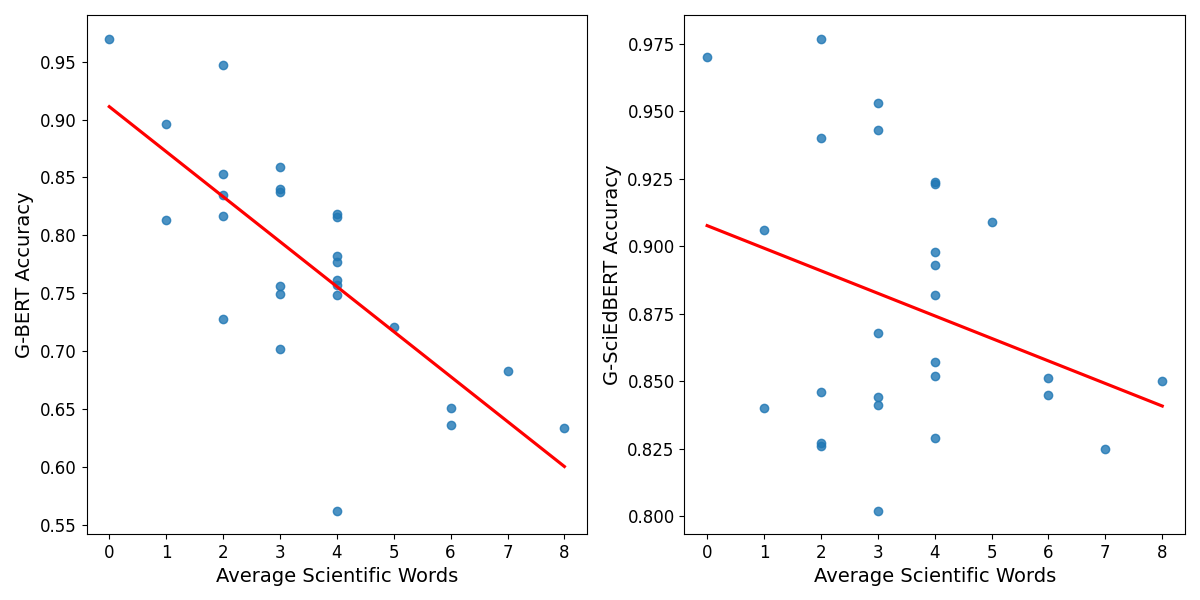}
        \caption{Effect of count of scientific words}
        \label{fig:scientific_words_vs_accuracy}
    \end{subfigure}
    \caption{Visualization of the effect of item-wise response length and the count of scientific words in student written responses with G-BERT and G-SciEdBERT accuracy. (Comparison plots for all 27 items). The less steep slope of the trend lines delineates the reduced effect of average sentence length and average scientific words against G-SciEdBERT accuracy.}
    \label{fig:combined_results}
\end{figure}

\subsection{Results}

Table \ref{tab:accuracyComparison} compares the outcomes between G-BERT and G-SciEdBERT for randomly picked five assessment items and the average accuracy (QWK) reported for all datasets combined. It shows that G-SciEdBERT significantly outperformed G-BERT in the automatic scoring of student-written responses. Based on the QWK values, the percentage differences in accuracy vary from 9.1\% to 14.1\%, with an average increase of 10.2\% (from .7758 to .8785). Especially item S131Q04 showed the greatest improvement of 14.1\% (from .683 to .825), which is noteworthy. These findings demonstrate that G-SciEdBERT is more effective than G-BERT at comprehending and assessing complex science-related writings. A comprehensive visualization of the accuracy results of all 27 items can be seen in Appendix \ref{apndx:table}.

To statistically validate these findings, we conducted a paired \textit{t}-test to compare the accuracy scores of G-BERT and G-SciEdBERT. The results revealed a significant difference in average accuracy between the two models ($t(26) = 29.24$, $p = 0.0017$), affirming the hypothesis that G-SciEdBERT holds a substantial advantage in interpreting and assessing scientific responses, likely due to its pre-training on science education material. The significance level of $p < 0.05$ underscores the reliability of these results, indicating that the observed differences are unlikely to be due to chance.

Further analysis was conducted to understand the influence of specific text characteristics, namely Average Sentence Length and Average Scientific Words, on the accuracy of these models. As depicted in Figure \ref{fig:combined_results}, G-BERT's accuracy showed a significant negative correlation with Average Sentence Length ($r = -0.72, F(1, 25) = 27.12, p < 0.001$), accounting for 52\% of the variance in accuracy. This trend suggests that G-BERT struggles more with longer sentences, possibly due to their inherent complexity and the model's limitations in handling extensive linguistic structures.

A similar, though less pronounced, pattern was observed with Average Scientific Words, where G-BERT's accuracy tended to decrease as the number of scientific terms increased. However, this relationship was not statistically significant, indicating that while there is an observable trend, it does not contribute as strongly to the model's performance variability as sentence length does.

In contrast, G-SciEdBERT demonstrated greater resilience to variations in both sentence length and scientific terminology, maintaining consistently higher accuracy levels. Although G-SciEdBERT also experienced a decrease in accuracy with longer sentences, the impact was less severe ($r = -0.31, F(1, 25) = 2.635, p = 0.117$), suggesting a better adaptation to complex linguistic structures. The influence of scientific words on G-SciEdBERT's accuracy was minimal and not statistically significant ($r = -0.32, F(1, 25) = 2.747, p = 0.110$), highlighting the model's robustness in processing specialized scientific content.

The study's findings offer strong assurance that can be applied towards the immediate aim of automatic scoring in educational systems, especially in science content-based testing. The enhanced performance demonstrated by G-SciEdBERT in the overall organization and at the level of specific linguistic characteristics proves the advantage of using science and education profiles over the general-purpose G-BERT models. Therefore, the outcomes of this study support the use of G-SciEdBERT in the large-scale learning assessments like PISA, where the accuracy of scoring is critical and the input text might contain domain-specific content.

\section{Discussion}

The results of this study underscore the efficacy of using domain-specific language models like G-SciEdBERT for automatic scoring in educational contexts. The significant improvement in scoring accuracy over the general-purpose G-BERT model, particularly in handling complex scientific texts, highlights the advantages of specialized training. G-SciEdBERT's ability to maintain higher accuracy across varying sentence lengths and the presence of scientific terminology demonstrates its robustness and suitability for science-oriented assessments.

Besides, it advances the development of automatic scoring and provides a practical reference for expanding such a model to other languages and areas. The working approach used here can be followed in other linguistic environments and academic disciplines, increasing the high accuracy and overall reliability of the systems for the automatic assessment of results worldwide.

For the research community interested in further exploration and application of G-SciEdBERT, the code repository is publicly available on GitHub (\url{https://github.com/ehsanlatif/G-SciEdBERT.git}), and the model itself can be accessed via Hugging Face (\url{https://huggingface.co/ai4stem-uga/G-SciEdBERT}). These resources enable the mirroring of this study and the contexts in which the model is adjusted for use in education.

Therefore, the outcomes of this study are rather important and pertinent for improving the methods of using specialized language models within the sphere of educational assessment and, in particular, for implementing such models in subjects where the assessment of context-sensitive and complex textual answers is of the utmost importance. The G-SciEdBERT study can be used as a reference for subsequent studies implementing the scheme of automatic scoring systems, enabling future advancements in deriving the highly needed beneficial outcomes for students across various educational contexts and subject areas.

\section{Conclusion}
In this study, we presented a contextualized model for the automatic scoring of German-written science responses, called G-SciEdBERT, which shows a significant accuracy gain over the general-purpose G-BERT model. By using a thorough analysis, German-SciEdBERT was able to increase its mean accuracy by about 10.2\% on a variety of PISA 2015 test items, demonstrating how well it comprehends and assesses complicated science-related argumentation. This important breakthrough demonstrates how specialized pre-training and fine-tuning can improve the accuracy of educational assessment models, establishing a new standard for AI applications in education and opening the door for further study into subject- and language-specific automatic scoring systems. G-SciEdBERT's outstanding performance is clearly attributed to its unique training and fine-tuning methods designed for the science education domain, indicating its promise as a dependable instrument in the global educational assessment environment.

%
%
%
\bibliographystyle{splncs04}
\bibliography{mybibliography}

\begin{thebibliography}{10}
\providecommand{\url}[1]{\texttt{#1}}
\providecommand{\urlprefix}{URL }
\providecommand{\doi}[1]{https://doi.org/#1}

\bibitem{achiam2023gpt}
Achiam, J., Adler, S., Agarwal, S., Ahmad, L., Akkaya, I., Aleman, F.L., Almeida, D., Altenschmidt, J., Altman, S., Anadkat, S., et~al.: Gpt-4 technical report. arXiv preprint arXiv:2303.08774  (2023)

\bibitem{alsanie2022automatic}
Alsanie, W., Alkanhal, M.I., Alhamadi, M., Alqabbany, A.O.: Automatic scoring of arabic essays over three linguistic levels. Progress in Artificial Intelligence pp. 1--13 (2022)

\bibitem{bressem2024medbert}
Bressem, K.K., Papaioannou, J.M., Grundmann, P., Borchert, F., Adams, L.C., Liu, L., Busch, F., Xu, L., Loyen, J.P., Niehues, S.M., et~al.: Medbert. de: A comprehensive german bert model for the medical domain. Expert Systems with Applications  \textbf{237},  121598 (2024)

\bibitem{chan2020german}
Chan, B., Schweter, S., M{\"o}ller, T.: German's next language model. arXiv preprint arXiv:2010.10906  (2020)

\bibitem{chen2023pmaes}
Chen, Y., Li, X.: Pmaes: Prompt-mapping contrastive learning for cross-prompt automated essay scoring. In: Proceedings of the 61st Annual Meeting of the Association for Computational Linguistics (Volume 1: Long Papers). pp. 1489--1503 (2023)

\bibitem{condor2021automatic}
Condor, A., Litster, M., Pardos, Z.: Automatic short answer grading with sbert on out-of-sample questions. International Educational Data Mining Society  (2021)

\bibitem{devlin2018bert}
Devlin, J., Chang, M.W., Lee, K., Toutanova, K.: Bert: Pre-training of deep bidirectional transformers for language understanding. arXiv preprint arXiv:1810.04805  (2018)

\bibitem{erickson2020automated}
Erickson, J.A., Botelho, A.F., McAteer, S., Varatharaj, A., Heffernan, N.T.: The automated grading of student open responses in mathematics. In: Proceedings of the tenth international conference on learning analytics \& knowledge. pp. 615--624 (2020)

\bibitem{fernandez2022automated}
Fernandez, N., Ghosh, A., Liu, N., Wang, Z., Choffin, B., Baraniuk, R., Lan, A.: Automated scoring for reading comprehension via in-context bert tuning. In: International Conference on Artificial Intelligence in Education. pp. 691--697. Springer (2022)

\bibitem{funayama2023can}
Funayama, H., Asazuma, Y., Matsubayashi, Y., Inui, T.M.K.: What can short answer scoring models learn from cross-prompt training data? In: Language Processing Society 29th Annual Conference (NLP2023), Okinawa. pp. 1874--1879 (2023)

\bibitem{funayama2023reducing}
Funayama, H., Asazuma, Y., Matsubayashi, Y., Mizumoto, T., Inui, K.: Reducing the cost: Cross-prompt pre-finetuning for short answer scoring. In: International conference on artificial intelligence in education. pp. 78--89. Springer (2023)

\bibitem{goodfellow2016deep}
Goodfellow, I., Bengio, Y., Courville, A.: Deep learning. MIT press (2016)

\bibitem{grassini2023shaping}
Grassini, S.: Shaping the future of education: exploring the potential and consequences of ai and chatgpt in educational settings. Education Sciences  \textbf{13}(7), ~692 (2023)

\bibitem{2023Gupta}
Gupta, A., Carpenter, D., Min, W., Rowe, J., Azevedo, R., Lester, J.: Detecting and mitigating encoded bias in deep learning-based stealth assessment models for reflection-enriched game-based learning environments. International Journal of Artificial Intelligence in Education pp. 1--28 (2023)

\bibitem{gururangan2020don}
Gururangan, S., Marasovi{\'c}, A., Swayamdipta, S., Lo, K., Beltagy, I., Downey, D., Smith, N.A.: Don't stop pretraining: Adapt language models to domains and tasks. arXiv preprint arXiv:2004.10964  (2020)

\bibitem{horbach2023crosslingual}
Horbach, A., Pehlke, J., Laarmann-Quante, R., Ding, Y.: Crosslingual content scoring in five languages using machine-translation and multilingual transformer models. International Journal of Artificial Intelligence in Education pp. 1--27 (2023)

\bibitem{iskender2020best}
Iskender, N., Polzehl, T., M{\"o}ller, S.: Best practices for crowd-based evaluation of german summarization: Comparing crowd, expert and automatic evaluation. In: Proceedings of the First Workshop on Evaluation and Comparison of NLP Systems. pp. 164--175 (2020)

\bibitem{kasneci2023chatgpt}
Kasneci, E., Se{\ss}ler, K., K{\"u}chemann, S., Bannert, M., Dementieva, D., Fischer, F., Gasser, U., Groh, G., G{\"u}nnemann, S., H{\"u}llermeier, E., et~al.: Chatgpt for good? on opportunities and challenges of large language models for education. Learning and individual differences  \textbf{103},  102274 (2023)

\bibitem{latif2023fine}
Latif, E., Zhai, X.: Fine-tuning chatgpt for automatic scoring. arXiv preprint arXiv:2310.10072  (2023)

\bibitem{lee2023gemini}
Lee, G.G., Latif, E., Shi, L., Zhai, X.: Gemini pro defeated by gpt-4v: Evidence from education. arXiv preprint arXiv:2401.08660  (2023)

\bibitem{2023Lee}
Lee, G.G., Latif, E., Wu, X., Liu, N., Zhai, X.: Applying large language models and chain-of-thought for automatic scoring (2023/11/30/ 2023), \url{http://arxiv.org/abs/2312.03748 files/12/2312.html}

\bibitem{liu2019roberta}
Liu, Y., Ott, M., Goyal, N., Du, J., Joshi, M., Chen, D., Levy, O., Lewis, M., Zettlemoyer, L., Stoyanov, V.: Roberta: A robustly optimized bert pretraining approach. arXiv preprint arXiv:1907.11692  (2019)

\bibitem{liu2023context}
Liu, Z., He, X., Liu, L., Liu, T., Zhai, X.: Context matters: A strategy to pre-train language model for science education. arXiv preprint arXiv:2301.12031  (2023)

\bibitem{ludwig2021automated}
Ludwig, S., Mayer, C., Hansen, C., Eilers, K., Brandt, S.: Automated essay scoring using transformer models. Psych  \textbf{3}(4),  897--915 (2021)

\bibitem{mayfield2020should}
Mayfield, E., Black, A.W.: Should you fine-tune bert for automated essay scoring? In: Proceedings of the Fifteenth Workshop on Innovative Use of NLP for Building Educational Applications. pp. 151--162 (2020)

\bibitem{nozza2020mask}
Nozza, D., Bianchi, F., Hovy, D.: What the [mask]? making sense of language-specific bert models. arXiv preprint arXiv:2003.02912  (2020)

\bibitem{pisa2015results}
{Organisation for Economic Co-operation and Development}: Pisa 2015 results (volume i): Excellence and equity in education. Tech. rep., OECD Publishing, Paris (2016). \doi{10.1787/9789264266490-en}, \url{https://doi.org/10.1787/9789264266490-en}

\bibitem{pisa2015techbook}
{Organisation for Economic Co-operation and Development}: Pisa 2015 technical report. Tech. rep., OECD Publishing, Paris (2017)

\bibitem{pisa2018results}
{Organisation for Economic Co-operation and Development}: Pisa 2018 results (volume i): What students know and can do. Tech. rep., OECD Publishing, Paris (2019). \doi{10.1787/5f07c754-en}, \url{https://doi.org/10.1787/5f07c754-en}

\bibitem{pisa2018techbook}
{Organisation for Economic Co-operation and Development}: Pisa 2018 technical report. Tech. rep., OECD Publishing, Paris (2019)

\bibitem{ridley2021automated}
Ridley, R., He, L., Dai, X.y., Huang, S., Chen, J.: Automated cross-prompt scoring of essay traits. In: Proceedings of the AAAI conference on artificial intelligence. vol.~35, pp. 13745--13753 (2021)

\bibitem{ridley2020prompt}
Ridley, R., He, L., Dai, X., Huang, S., Chen, J.: Prompt agnostic essay scorer: a domain generalization approach to cross-prompt automated essay scoring. arXiv preprint arXiv:2008.01441  (2020)

\bibitem{sasaki2022sentence}
Sasaki, T., Masada, T.: Sentence-bert distinguishes good and bad essays in cross-prompt automated essay scoring. In: 2022 IEEE International Conference on Data Mining Workshops (ICDMW). pp. 274--281. IEEE Computer Society (2022)

\bibitem{sawatzki2021deep}
Sawatzki, J., Schlippe, T., Benner-Wickner, M.: Deep learning techniques for automatic short answer grading: Predicting scores for english and german answers. In: International Conference on Artificial Intelligence in Education Technology. pp. 65--75. Springer (2021)

\bibitem{scheible2020gottbert}
Scheible, R., Thomczyk, F., Tippmann, P., Jaravine, V., Boeker, M.: Gottbert: a pure german language model. arXiv preprint arXiv:2012.02110  (2020)

\bibitem{shen2021mathbert}
Shen, J.T., Yamashita, M., Prihar, E., Heffernan, N., Wu, X., Graff, B., Lee, D.: Mathbert: A pre-trained language model for general nlp tasks in mathematics education. arXiv preprint arXiv:2106.07340  (2021)

\bibitem{song2020multi}
Song, W., Zhang, K., Fu, R., Liu, L., Liu, T., Cheng, M.: Multi-stage pre-training for automated chinese essay scoring. In: Proceedings of the 2020 Conference on Empirical Methods in Natural Language Processing (EMNLP). pp. 6723--6733 (2020)

\bibitem{wolf2020transformers}
Wolf, T., Debut, L., Sanh, V., Chaumond, J., Delangue, C., Moi, A., Cistac, P., Rault, T., Louf, R., Funtowicz, M., et~al.: Transformers: State-of-the-art natural language processing. In: Proceedings of the 2020 conference on empirical methods in natural language processing: system demonstrations. pp. 38--45 (2020)

\bibitem{2023Wu}
Wu, X., He, X., Li, T., Liu, N., Zhai, X.: Matching Exemplar as Next Sentence Prediction (MeNSP): Zero-shot Prompt Learning for Automatic Scoring in Science Education, vol. LNAI 13916, p. 1–13. Springer, Switzerland AG (2023). \doi{https://doi.org/10.1007/978-3-031-36272-9\_33}

\bibitem{zhai2022applying}
Zhai, X., He, P., Krajcik, J.: Applying machine learning to automatically assess scientific models. Journal of Research in Science Teaching  \textbf{59}(10),  1765--1794 (2022)

\bibitem{zhai2020applying}
Zhai, X., Yin, Y., Pellegrino, J.W., Haudek, K.C., Shi, L.: Applying machine learning in science assessment: a systematic review. Studies in Science Education  \textbf{56}(1),  111--151 (2020)

\end{thebibliography}

\appendix
\section{Accuracy Results for all 27 items}
\label{apndx:table}

\begin{table}[h]
\centering
\caption{Comparison between G-BERT and G-SciEdBERT on automatic scoring of 27 PISA 2015 science items excluding responses of items from PISA 2018.}
\label{tab:accuracyComparison}
\resizebox{\linewidth}{!}{
\begin{tabular}{@{}lcccccc@{}}
\toprule
& \multicolumn{2}{c}{ Samples (\textit{n}) } & \multicolumn{2}{c}{Accuracy as QWK} \\
Item & Training Samples & Testing Samples & G-BERT & \textbf{G-SciEdBERT} \\ \midrule
S131Q02 & 487 & 122 & 0.761 & 0.852 \\
S131Q04 & 478 & 120 & 0.683 & 0.825 \\
S268Q02 & 446 & 112 & 0.757 & 0.893 \\
S269Q01 & 508 & 127 & 0.837 & 0.953 \\
S269Q03 & 500 & 126 & 0.702 & 0.802 \\
S304Q01 & 480 & 120 & 0.633 & 0.850 \\
S438Q03 & 433 & 109 & 0.817 & 0.826 \\
S458Q01 & 506 & 127 & 0.782 & 0.898 \\
S465Q01 & 516 & 129 & 0.636 & 0.845 \\
S495Q03 & 475 & 119 & 0.816 & 0.924 \\
S498Q04 & 423 & 106 & 0.840 & 0.868 \\
S510Q04 & 491 & 123 & 0.859 & 0.943 \\
S514Q02 & 525 & 132 & 0.947 & 0.977 \\
S514Q03 & 464 & 117 & 0.835 & 0.846 \\
S514Q04 & 526 & 132 & 0.970 & 0.970 \\
S519Q01 & 525 & 132 & 0.756 & 0.841 \\
S519Q03 & 508 & 127 & 0.896 & 0.906 \\
S524Q07 & 381 & 96 & 0.813 & 0.840 \\
S602Q03 & 713 & 179 & 0.749 & 0.844 \\
S603Q02 & 814 & 204 & 0.777 & 0.882 \\
S604Q04 & 872 & 219 & 0.721 & 0.909 \\
S605Q04 & 830 & 208 & 0.818 & 0.923 \\
S607Q03 & 859 & 215 & 0.728 & 0.827 \\
S648Q05 & 779 & 195 & 0.651 & 0.851 \\
S649Q02 & 935 & 234 & 0.853 & 0.940 \\
S656Q02 & 889 & 223 & 0.748 & 0.857 \\
S657Q04 & 776 & 194 & 0.562 & 0.829 \\
\bottomrule
\end{tabular}
}
\end{table}

\end{document}